# Using Machine Learning to Predict Air Quality Index in New Delhi

Samayan Bhattacharya
Department of Computer Science and Engineering
Jadavpur University
samayan.bhattacharya@gmail.com

Sk Shahnawaz
Department of Computer Science and Engineering
Jadavpur University
skshahnawaz2909@gmail.com

*Abstract*—Air quality has a significant impact on human health. Degradation in air quality leads to a wide range of health issues, especially in children. The ability to predict air quality enables the government and other concerned organizations to take necessary steps to shield the most vulnerable, from being exposed to the air with hazardous quality. Traditional approaches to this task have very limited success because of a lack of access of such methods to sufficient longitudinal data. In this paper, we use a Support Vector Regression (SVR) model to forecast the levels of various pollutants and the air quality index, using archive pollution data made publicly available by Central Pollution Control Board and the US Embassy in New Delhi. Among the tested methods, a Radial Basis Function (RBF) kernel produced the best results with SVR. According to our experiments, using the whole range of available variables produced better results than using features selected by principal component analysis. The model predicts levels of various pollutants, like, sulfur dioxide, carbon monoxide, nitrogen dioxide, particulate matter 2.5, and ground-level ozone, as well as the Air Quality Index (AQI), at an accuracy of 93.4 percent.

## I. Introduction

The sharp rise in air pollution in recent years, due to industrial and agricultural activities, as well as increased number of vehicles using internal combustion engines, has caught the attention of the scientific community [1, 2, 3]. Air pollution has significant impact on human health and may cause long-term health issues in children. The significant rise in air pollution in New Delhi is attributed to increased vehicular emissions, burning of fossil fuels at power plants, and other local industries and burning of fields by farmers in neighboring states [4].

Air quality is being monitored in New Delhi for about two decades. This has allowed a better understanding of the changes in air pollution in response to particular activities and government regulations, but the air pollution in New Delhi remains a problem [5].

Air pollution is responsible for 30 percent of lower-respiratory tract infections and is linked with 91percent of premature deaths, from lung cancer, heart disease, acute respiratory infections, stroke and chronic obstructive pulmonary disease. It contributes to 20 percent of infant mortality worldwide and causes numerous short- and long-term illnesses in children. Exposure of the mother to high levels of air pollution can lead to adversely affect immune status, brain development, respiratory systems, and cardio-metabolic health of the child. Air pollution has also been linked to low birth weight and stunted growth in children.

Air pollution is estimated to be responsible for one in ten deaths of children under five years of age. In elder people, air pollution causes high rates of asthma, with decreased cognitive performance.

Presently, the government implements regulations after the air quality reaches hazardous levels. If there is a way to foresee the air quality reaching hazardous levels, the government can implement such regulations early, potentially preventing further degradation of air quality and being able to shield those, most vulnerable, from getting exposed to such air quality. This study aims to build a model that can look at previously recorded air quality data and predicts levels of different pollutants as well as air quality index. For this we use a variation of Support Vector Machines (SVM), called Support Vector Regression (SVR).

The paper is organized as follows. We state the motivations of this work and frame our work in section 2, stating the potential impact of being able to successfully predict the air quality. We provide a critical revision of related work, done previously, in Section 3. We explain how Support Vector Machines (SVM), particularly Support Vector Regression works in Section 4. We describe the datasets used in this work and the data preprocessing steps used to produce a more efficient input for the SVR, in sections 5 and 6 respectively. In Section 7, we present details of the experiment performed, divided into subsections describing the experimental setup and the results obtained. In Section 8, we conclude the paper and discuss ideas for future work in this area.

## II. Background and Motivation

Air pollution is caused by the introduction of harmful or excessive quantities of certain substances into the atmosphere. Such substances include solid particles, liquid droplets and gases. Air pollutants are classifieds into primary and secondary pollutants. Primary pollutants are the pollutants that are directly released from their source directly into the atmosphere. Sources of primary air pollutants may be natural, like volcanic eruptions, sand storms, etc. or man-made, like burning of fossil fuels, leaking gases from appliances, etc. Primary pollutants include sulfur dioxide ($SO_2$), oxides of nitrogen ($NO_x$), particulate matter (PM) and carbon monoxide (CO). Secondary pollutants are formed in the atmosphere due to chemical or physical interactions between primary pollutants. Secondary pollutants include photochemical oxidants and secondary particulate matter.

The most common pollutants are called criteria pollutants and correspond to the most prevalent health threats. These include $SO_2$, ground level ozone ($O_3$), $NO_2$, lead and PM. It has been demonstrated that there is a correlation between exposure to such pollutants for a short while and health issue like inflamed respiratory track in healthy people, increased respiratory symptoms in people with asthma, difficulty meeting high oxygen requirements while exercising and critical respiratory situations, especially in children and the elderly [6].

National agencies like EPA, EU, and many others have set standards for acceptable levels of air quality. Air quality index (AQI) is used to indicate the levels of the criteria pollutants in the air. The overall AQI is the maximum of the AQI recorded for the individual criteria pollutants. AQI levels also indicate the health risks associated with exposure to the particular air quality. Such health symptoms may be experienced shortly after exposure to polluted air or in the long run. Such symptoms may also vary based on the age and health conditions of the particular person being exposed.

Thus, it is vital that we have a system to forecast increases in air pollution levels, so that government organizations may be able to counter further increase through on-demand pollution control mechanisms or an emergency response [7]. This would make AQI more controllable to suit the overall needs of the population. If the rise in pollution cannot be curbed, the authorities may issue warning to the population about the AQI forecast in order to shield the most vulnerable from getting exposed in case the AQI forecast exceeds the permitted limit.

### III. Previous and Related Work

The best statistical method for predicting time series data is the autoregressive integrated moving average model (ARIMA) [8]. It has several advantages in terms of its statistical properties [9], potential for a wide range of applications and extendibility. With the rise in importance for predicting air quality levels, ARIMA was applied to this task as well. It was demonstrated to reach accuracies around 95% [10] for forecasting AQI monthly values. [11] compared the performance of ARIMA with a Holt exponential smoothing model and proved the superiority of the ARIMA model for forecasting AQI daily values. However, this method requires extensive manual intervention in terms of selecting the data fed into the system, as it has a low tolerance towards outliers. The features to be considered must also be selected.

The availability of large quantities of archive data made it convenient to use Machine Learning (ML) [12] models for the time series prediction of AQI. ML models are able to automatically look at large amounts of data and select important features, thus reducing the need for human intervention. ML models are able to achieve higher accuracies with large datasets, than classic statistical methods. Such models have long been used for AQI forecasting tasks.

ML models are nonlinear, nonparametric in nature and hence are better able to handle the complexity of nonlinear elements like pollutant levels in the air [13]. Hence they outperform statistical methods like ARIMA, Winter exponential smoothing, and multivariate regression, which work well only with linear systems [14, 15]. ML models like Artificial Neural Networks (ANN), Genetic programming and Support Vector Machine (SVM) are able to find hidden patterns in vast quantities of data [7].

Several works have used Support Vector Machines (SVM) for predicting time series data. Some works have also used SVM for forecasting air quality. [16] proposed the use a variant of SVM for regression tasks and called it Support Vector Regression (SVR). [17] proved the superiority of SVR over Artificial Neural Networks (ANN). [18] showed that a hybrid of ANN and SVM produced better results. They used ANN for partitioning the input space and the SVM to model each portion.

Some of the works that used SVM for time series air quality forecasting include, (1) the model for the prediction of air quality in downtown Hong Kong by [19], showing that SVMs perform better than other Machine Learning approaches, (2) SVM model by [20] for air quality (PM10) forecasting in Bangkok, (3) SVM for air quality forecasting in Macau by [21] (4) work of [22] for forecasting in Mexico City, that lead to the conclusion that SVMs are more scalable and flexible for nonlinear, dynamic data, (5) the hybrid model proposed by [23], combing the advantages of SVM and flower pollination algorithm, which was shown to outperform any particular model.

### IV. Support Vector Machine

Support Vector Machines (SVM) were introduced by [25] as a classification technique. The objective is to use the hyperplane to separate the data, represented as support vectors, belonging to the different classes. When the original data is not linearly separable, it is projected to a higher dimension, using a kernel function. This makes the nonlinearly separable data linearly separable.

Support Vector Regression (SVR) was introduced by [24]. This allowed Support Vector Machines to be applied to regression using a new loss function. SVR has been used for time series forecasting by several works [16, 17, 26]. It has been demonstrated that SVR models offer faster training and better forecast ability while using smaller number of parameters.

The objective of SVR model is to learn a nonlinear mapping $\varphi_j(X)$ of the data to a high dimensional vector space such that the projections can be used to train a linear regression model. The trained linear regression model is then used to forecast in the high dimensional space after mapping the input to the high dimensional space using the kernel function.

The SVR model uses a combination of the training error and a regularization term in the loss function. Apart from this, other interesting properties arise from the use of kernel function, enabling it to be used for both linear and nonlinear forecasting and the convex nature of the fitness function and its constraints.

Let, the training set with m data points be represented as

$$T=\{(x_1,y_1),(x_2,y_2),(x_3,y_3),(x_4,y_4)\ldots(x_m,y_m)\} \quad (1)$$

where, $x \in X \subset R^n$ are the inputs in the training set and $y \in Y \subset R$ are the corresponding expected outputs in the training set.

A nonlinear kernel function is represented as

$$f(x) = \omega^T \Phi(x_i) + b \quad (2)$$

Equation (2) can be written in the form of a constrained convex optimization problem as:

minimize $\frac{1}{2}\omega^T\omega$

(3)

subject to
$$\begin{cases} y_i - \omega^T \Phi(x_i) - b \leq \varepsilon \\ \omega^T \Phi(x_i) + b - y_i \leq \varepsilon \end{cases}$$

The aim of this objective function is to minimize ω, while satisfying the other constraints, under the assumption that convex optimization problem is feasible, that is, f(x) exists. In case this assumption is not true, errors can be traded off for the flatness of the estimate. Thus, (3) can be reformulated as (4).

$$\text{minimize} \quad \frac{1}{2}\omega^T \omega + C \sum_{i=1}^{m}(\epsilon_i^+ + \epsilon_i^-)$$

(4)

subject to
$$\begin{cases} y_i - \omega^T \Phi(x_i) - b \leq \varepsilon + \epsilon_i^+ \\ \omega^T \Phi(x_i) + b - y_i \leq \varepsilon + \epsilon_i^- \\ \epsilon_i^+ \epsilon_i^- \geq 0 \end{cases}$$

where C<0 represents weights of the loss function and is an initialized constant. $\omega^T \omega$ is the regularized term and $C \sum_{i=1}^{m}(\epsilon_i^+ + \epsilon_i^-)$ is the empirical term, measuring the ε-insensitive loss function.

While solving (4) Lagragian multipliers $(\alpha_i^+, \eta_i^+, \alpha_i^-, \eta_i^-)$ may be used to eliminate some of the primal variables. The final equation translating the dual optimization problem is (5).

$$\text{minimize} \quad \frac{1}{2}\sum_{i,j=1}^{m} K(x_i, x_j)(\alpha_i^+ - \alpha_i^-)(\alpha_j^+ - \alpha_j^-) + \varepsilon \sum_{i=1}^{m}(\alpha_i^+ + \alpha_i^-) - \sum_{i=1}^{m}(\alpha_i^+ - \alpha_i^-)$$

(5)

subject to $\sum_{i=1}^{m}(\alpha_i^+ - \alpha_i^-) = 0$
$\alpha_i^+, \alpha_i^- \in [0, C]$

where K($x_i$, $x_j$) represents the kernel function, allowing the application of SVR to nonlinear functions. The performance of the SVR model depends on the kernel function, the regularization parameter (C) and the insensitive parameter (ε). There are many options for the kernel function [27]. In this work we study radial basis function (RBF) and polynomial kernel.

## V. Data Description

The datasets used in this study were obtained from the archive data provided by the US Embassy in Delhi and the Central Pollution Control Board. These datasets are described below.

The data from the US Mission in India consists of hourly concentrations of particulate matter of sizes less than or equal to 2.5 microns (PM2.5) and particulate matter of diameter less than or equal to 10 microns (PM10). They derive the Air Quality Index based on these values. The values are recorded using device located on the campus of the embassy and are thus, highly local and different from those recorded by the Central Pollution Control Board.

The data from Central Pollution Control Board are daily recordings of the concentrations of Sulfur dioxide ($SO_2$), Nitrogen dioxide ($NO_2$), PM2.5, PM10, and suspended particulate matter (SPM). The data is of relatively poor quality with significant amount of missing values. The recordings are from 4 recording stations in different parts of Delhi but are reasonably different from one another. Since the data is recorded daily instead of hourly, the continuity of the time series representation of the data is adversely affected and must be handled in the pre-processing step.

## VI. Data Preprocessing

Data quality and effective data representation are of paramount importance in ensuring good performance of a forecasting model and its generalizability [28] of the SPM data. The standard data processing steps include (1) preparing more accurate and complete datasets by imputing missing data and removing or modifying outlier data points, (2) ensuring data is uniformly distributed by normalization and standardization of data, (3) creating a smaller and compact dataset by extraction and selection of features. We perform these steps on our data as follows.

**Imputation of missing data:** We found that more than 50% data was missing, so we removed the SPM field from the Central Pollution Control Board data. For the other fields in the US embassy data and the Central Pollution Control Board data, we substituted missing data by second order polynomial estimation using nearest available data points. It gave better results than using series mean or linear interpolation.

**Removing or modifying outliers:** An irregular pattern was observed in pollutant data between August and October 2020 in both US embassy data and the Central Pollution Control Board data. Thus, these data were removed. For data modification, we used the power transformation method [29]. This provides a nonlinear transformation that is more robust to noise and hence produces better data.

**Feature extraction:** The date component in the Central Pollution Control Board data and the date-time component in the US embassy data were used to produce new features. The date component was used to obtain a field called seasons. Four seasons were used (Summer, Fall, Winter, Spring). The cyclic nature of the time component was exploited to obtain two fields {sin(2πhour/24), cos(2πhour/24)}. The date component was also used to obtained fields for day, month and year.

**Feature selection:** From the features obtained in the previous step using feature engineering, a few variables were selected to reduce dimensionality of the dataset and remove collinearity. Correlation-based feature selection was used [30], to check for collinearity among features. It was observed that the concentration of some of the pollutants had an almost linear correlation. For example, NO2 and CO concentrations were almost linearly related and so were CO and PM2.5 concentrations. Based on the remarkable correlation [31], it was decided to keep all pollutants in the dataset. In spite of SVR models being robust against collinearity and multicollinearity [32], we dropped some variables showing strong correlation n with some other variable. For example, the season variable had strong correlation with the month variable, hence we dropped the month variable, also, the hour variable was dropped due to strong correlation with the hour sin and hour cos cyclic variables. We also used Principal Component Analysis (PCA) [33] for reducing the dimensionality of the dataset. It enabled us to reduce the number of variables for each pollutant by about 76%. We compare the results with and without PCA in the next section.

VII. EXPERIMENTAL STUDY

Experimental settings: There are three user defined hyperparameters in an SVR model, the maximum allowed deviation ε, the regularization constant C, and the kernel type function. For determining C and ε, time-series split, combined with random grid search was employed [34, 35]. The range of C was extended to be 1 to 100, to allow wider exploration, as opposed to the 10 to 100 range [26]. The range of ε was taken to be between 0.001 and 0.1, with a step of 0.001. The most popular kernel functions being RBF and polynomial, we compare results for both of them. The optimum number of iterations for the random search was taken to be 60 [36].

**Experimental results:** Here we discuss the performance of the SVR models in forecasting the levels of 4 pollutants, nitrogen dioxide ($NO_2$), sulfur dioxide ($SO_2$), particulate matter 2.5 ($PM_{2.5}$) and particulate matter 10 ($PM_{10}$) with and without Principal Component Analysis (PCA).

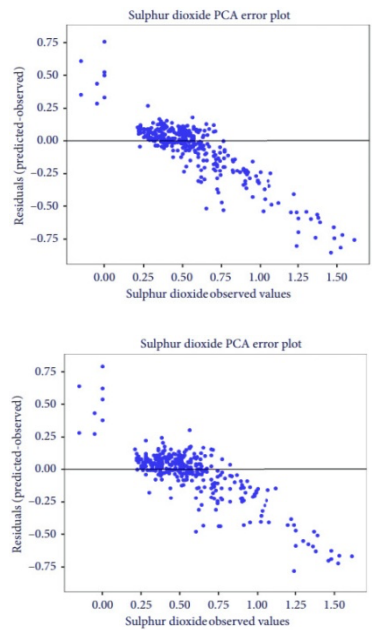

Figure 2: (a) PCA SVR-RBF forecasting errors plotted against observed SO2 values and (b) SVR-RBF forecasting errors plotted against observed SO2 values.

TABLE II. ERROR METRICS OF THE FORECASTING MODEL FOR $SO_2$ LEVEL DETECTION

| Error Metrics | PCA SVR-RBF | | SVR-RBF | |
|---|---|---|---|---|
| | Training Set | Validation Set | Training Set | Validation Set |
| MAE | 0.105 | 0.228 | 0.094 | 0.161 |
| $R^2$ | 0.974 | 0.884 | 0.980 | 0.938 |
| RMSE | 0.151 | 0.315 | 0.133 | 0.237 |
| nRMSE | 0.028 | 0.050 | 0.024 | 0.037 |

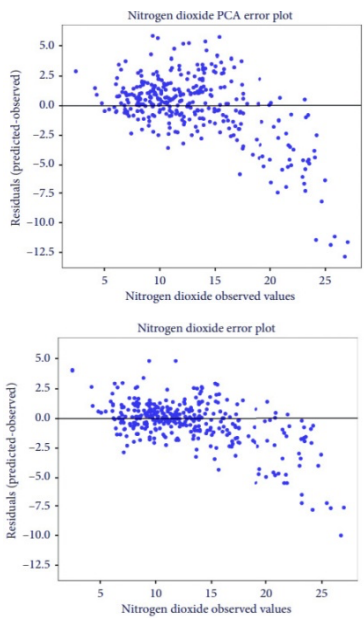

Figure 1: (a) PCA SVR-RBF forecasting errors plotted against observed NO2 values and (b) SVR-RBF forecasting errors plotted against observed NO2 values

TABLE I. ERROR METRICS OF THE FORECASTING MODEL FOR $NO_2$ LEVEL DETECTION

| Error Metrics | PCA SVR-RBF | | SVR-RBF | |
|---|---|---|---|---|
| | Training Set | Validation Set | Training Set | Validation Set |
| MAE | 0.235 | 0.460 | 0.228 | 0.413 |
| $R^2$ | 0.788 | 0.024 | 0.831 | 0.272 |
| RMSE | 0.363 | 0.753 | 0.351 | 0.702 |
| nRMSE | 0.037 | 0.054 | 0.034 | 0.048 |

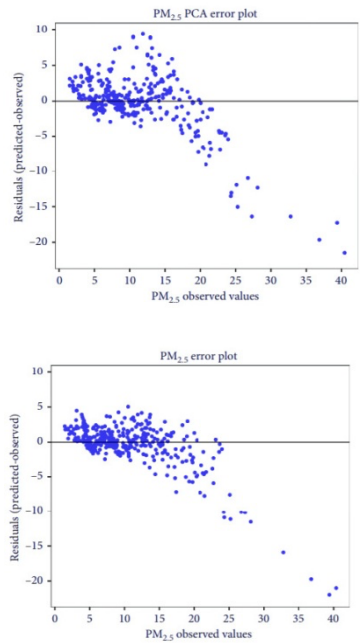

Figure 3: (a) PCA SVR-RBF forecasting errors plotted against observed PM2.5 values and (b) SVR-RBF forecasting errors plotted against observed PM2.5 values.

| TABLE III. | ERROR METRICS OF THE FORECASTING MODEL FOR $PM_{2.5}$ LEVEL DETECTION |

| Error Metrics | PCA SVR-RBF | | SVR-RBF | |
|---|---|---|---|---|
| | Training Set | Validation Set | Training Set | Validation Set |
| MAE | 0.201 | 0.381 | 0.145 | 0.330 |
| $R^2$ | 0.881 | 0.562 | 0.936 | 0.646 |
| RMSE | 0.274 | 0.577 | 0.204 | 0.511 |
| nRMSE | 0.511 | 0.073 | 0.030 | 0.067 |

| TABLE IV. | ERROR METRICS OF THE FORECASTING MODEL FOR $PM_{10}$ LEVEL DETECTION |

| Error Metrics | PCA SVR-RBF | | SVR-RBF | |
|---|---|---|---|---|
| | Training Set | Validation Set | Training Set | Validation Set |
| MAE | 0.084 | 0.191 | 0.040 | 0.089 |
| $R^2$ | 0.988 | 0.922 | 0.997 | 0.981 |
| RMSE | 0.111 | 0.261 | 0.061 | 0.134 |
| nRMSE | 0.024 | 0.051 | 0.012 | 0.024 |

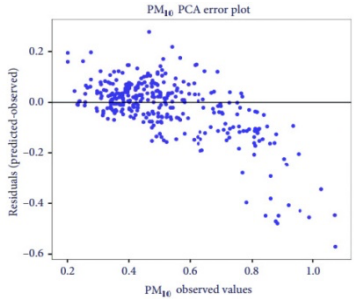
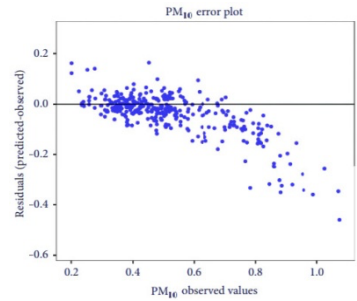

Figure 4: (a) PCA SVR-RBF forecasting errors plotted against observed PM10 values and (b) SVR-RBF forecasting errors plotted against observed PM10 values.

Figure 5: Six AQI categories defined by EPA

TABLE V. CONFUSION MATRIX FOR THE AQI CLASSIFICATIONS OBTAINED WITH BOTH MODELS FOR THE TRAINING SET

| Training Dataset | PCA SVR-RBF | | | SVR-RBF | | |
|---|---|---|---|---|---|---|
| | *Good* | *Moderate* | *Unhealthy* | *Good* | *Moderate* | *Unhealthy* |
| Good | 2509 | 159 | 0 | 2547 | 106 | 0 |
| Moderate | 306 | 992 | 0 | 252 | 1045 | 0 |
| Unhealthy | 6 | 24 | 0 | 9 | 19 | 0 |

TABLE VI. CONFUSION MATRIX FOR THE AQI CLASSIFICATIONS OBTAINED WITH BOTH MODELS FOR THE VALIDATION SET

| Training Dataset | PCA SVR-RBF | | | SVR-RBF | | |
|---|---|---|---|---|---|---|
| | *Good* | *Moderate* | *Unhealthy* | *Good* | *Moderate* | *Unhealthy* |
| Good | 1166 | 54 | 0 | 1179 | 45 | 0 |
| Moderate | 61 | 425 | 0 | 49 | 439 | 0 |
| Unhealthy | 0 | 2 | 0 | 9 | 3 | 0 |

## VIII. CONCLUSION

The task of forecasting pollutant levels is inherently hard because of the volatile and dynamic nature of the data and its variability in space and time. However, the task of forecasting pollutant levels has been increasing in importance due to the effects of pollution on the population and the environment. In this work we used SVR for forecasting levels of pollutants like $NO_2$, $SO_2$, $PM_{2.5}$ and $PM_{10}$, and Air Quality Index (AQI), using publicly available data for New Delhi.

As the next step, we would like to investigate and compare the performance of other Machine Learning methods like Artificial Neural Network (ANN) and genetic algorithms, for this task. We would also like to explore the use other methods of hyperparameter optimization and other methods of variable selection for larger datasets.

ACKNOWLEDGMENT

The authors would like to acknowledge the support of the Central Pollution Control Board (CBCP) and the Government of Delhi in collecting the data for this project